\documentclass{article}
\usepackage{ijcai20}

\usepackage{times}
\usepackage{soul}
\usepackage{url}
\usepackage[hidelinks]{hyperref}
\usepackage[utf8]{inputenc}
\usepackage[small]{caption}
\usepackage{subcaption}
\usepackage{graphicx}
\usepackage{amsmath}
\usepackage{dsfont}
\usepackage{amsthm}
\usepackage{booktabs}
\usepackage{algorithm}
\usepackage{algorithmic}
\usepackage{xcolor}
\usepackage{multirow}
\usepackage[normalem]{ulem}
\useunder{\uline}{\ul}{}

\newcommand{\paperTitle}{Small, Accurate, and Fast Vehicle Re-ID on the Edge: the SAFR Approach}

\newcommand{\safr}{\texttt{SAFR}}

\newcommand{\slarge}{\texttt{SAFR-Large}}
\newcommand{\smed}{\texttt{SAFR-Medium}}
\newcommand{\smobile}{\texttt{SAFR-Small}}
\newcommand{\sedge}{\texttt{SAFR-Micro}}

\newcommand{\largemap}{81.34}
\newcommand{\largerank}{96.93}

\newcommand{\medmap}{79.34}
\newcommand{\medrank}{93.34}
\newcommand{\medmapdrop}{2.5\%}
\newcommand{\medrankdrop}{2.7\%}
\newcommand{\medsmall}{30\%}
\newcommand{\medspeed}{18\%}

\newcommand{\mobilemap}{77.14}
\newcommand{\mobilerank}{93.14}
\newcommand{\mobilemapdrop}{5.2\%}

\newcommand{\mobilesmall}{60\%}
\newcommand{\mobilespeed}{150\%}

\newcommand{\edgemap}{75.80}
\newcommand{\edgerank}{92.61}
\newcommand{\edgemapdrop}{6.8\%}
\newcommand{\edgerankdrop}{4.5\%}
\newcommand{\edgesmall}{95\%} 
\newcommand{\edgespeed}{3362\%}

\newcommand{\resnetlarge}{\texttt{ResNet-50}}
\newcommand{\resnetsmall}{\texttt{ResNet-18}}
\newcommand{\resnetmed}{\texttt{ResNet-34}}
\newcommand{\shufflenet}{\texttt{ShuffleNet-v2+}}

\DeclareMathOperator{\sgn}{sgn}

\def\Snospace~{\color{red}\S{}}

\newcommand{\squishitemize}{
\begin{list}{$\bullet$}
	{ \setlength{\itemsep}{0pt}
		\setlength{\parsep}{3pt}
		\setlength{\topsep}{3pt}
		\setlength{\partopsep}{0pt}
		\setlength{\leftmargin}{1.95em}
		\setlength{\labelwidth}{1.5em}
		\setlength{\labelsep}{0.5em} } }

\newcounter{Lcount}
\newcommand{\squishlist}{
\begin{list}{\arabic{Lcount}. }
	{ \usecounter{Lcount}
		\setlength{\itemsep}{0pt}
		\setlength{\parsep}{3pt}
		\setlength{\topsep}{3pt}
		\setlength{\partopsep}{0pt}
		\setlength{\leftmargin}{2em}
		\setlength{\labelwidth}{1.5em}
		\setlength{\labelsep}{0.5em} } }

\newcommand{\squishend}{\end{list}}

\newcommand{\PP}[1]{
	\vspace{2px}
	\noindent{\bf {#1}{.}}
}

\title{\paperTitle}

\author{
    Abhijit Suprem$^1$, Calton Pu$^1$, Joao Eduardo Ferreira$^2$ 
    \affiliations
    $^1$School of Computer Science, Georgia Institute of Technology, USA
    $^2$University of Sao Paulo, Brazil
    \emails
	asuprem@gatech.edu
}

\begin{document}

\maketitle
\begin{abstract}


We propose a Small, Accurate, and Fast Re-ID (\safr{}) design for flexible vehicle re-id under a variety of compute environments such as cloud, mobile, edge, or embedded devices by only changing the re-id model backbone. Through best-fit design choices, feature extraction, training tricks, global attention, and local attention, we create a re-id model design that optimizes multi-dimensionally along model size, speed, \& accuracy  for deployment under various memory and compute constraints. We present several variations of our flexible \safr{} model: \slarge{} for cloud-type environments with large compute resources, \smobile{}  for mobile devices with some compute constraints, and \sedge{} for edge devices with severe memory \& compute constraints. 

\slarge{} delivers state-of-the-art results with mAP \largemap{} on the VeRi-776  vehicle re-id dataset (15\% better than related work). \smobile{} trades a \mobilemapdrop{} drop in performance (mAP \mobilemap{} on VeRi-776) for over \mobilesmall{} model compression and \mobilespeed{} speedup. \sedge{}, at only 6MB and 130MFLOPS, trades \edgemapdrop{} drop in accuracy (mAP \edgemap{} on VeRi-776) for \edgesmall{} compression and 33x speedup compared to \slarge{}.

\end{abstract}

\section{Introduction}
\label{sec:intro}
Increasing numbers of traffic camera networks in the wild have coincided with growing attention towards the traffic management problem, where the goal is to improve vehicle detection and recognition for better traffic safety and emergency response. Several works in the past decade have focused on this problem \cite{killer,killer2}. An important consideration is small and fast models for edge and mobile deployment on cameras to reduce bandwidth costs and distribute processing from cloud to edge \cite{killer}. The primary challenge remains vehicle re-id where the same vehicle must be identified across multiple cameras \cite{killer3}.

\PP{Vehicle Re-ID} There have been several recent works towards accurate vehicle re-id \cite{ram,oife,vami,ealn,gstre}. Vehicle re-id has two primary challenges: (i) inter-class similarity, where two different vehicles appear similar due to assembly-line manufacturing, and (ii) intra-class variability, where the same vehicle looks different due to different camera orientations. Re-id relies on representational learning, where a model learns to track vehicles by learning fine-grained attributes such as decals or emblem. End-to-end re-id models using automatic feature extraction for re-id are common; these approaches are  designed for offline  re-id since due to their complexity, size, and compute cost, they  are suitable only for cloud environments with more compute resources.

\PP{Multi-dimensional Models} For vehicle re-id in video data from traffic monitoring cameras, there is an opportunity for flexible and scalable models running on the cloud, edge, or mobile devices, depending on varied performance (e.g., real-time) or resource requirements \cite{chameleon}: edge re-id can be used for local tracking, mobile re-id can be used for validating edge re-id results, and cloud re-id can be used for sophisticated and complex vehicle re-id models. Theoretically, different models could be used in different system tiers, but it would be expensive to develop and maintain those disparate models, and their non-trivial interactions may degrade system performance. In this paper, we propose the Small, Accurate, and Fast Re-id (\safr{}) approach, capable of generating classification models with significantly different sizes and speeds (from 0.18GFLOPS to 4GFLOPS), while preserving model accuracy (with less than 10\% loss from the largest to smallest of models). The SAFR approach to vehicle re-id enables flexible and effective resource management from cloud to edge, without requiring major changes or integration of different models, simplifying deployability and improving monitoring efficiency \cite{flexibility}.


\PP{The \safr{} Approach} In this paper, we propose \safr{} - a small, accurate, and fast re-id design that achieves state-of-the-art performance on vehicle re-id across a variety of datasets. Since vehicle re-id requires identifying fine-grained local attributes of vehicles across camera orientations, we develop an unsupervised parts-based local features extractor to detect vehicle parts across orientations. By learning fine-grained variances between vehicles, e.g. headlights, decals, emblem, \safr{}'s local attention modules for local feature extraction addresses the intra-class variability problem. Simultaneously, \safr{} uses a global attention module to ensure that models do not overfit on fine-grained features, thereby retaining important contextual information. \safr{} performs rich feature extraction on a single backbone compared to multi-branch networks with expensive fully connected layers covered in \autoref{sec:related}, allowing re-id models based on \safr{}'s design to be both smaller and faster.

\PP{Contributions} First, we present \safr{}, a small, accurate, and fast multi-dimensional model design approach for vehicle re-id that integrates global attention, local attention, ground-up backbone design, and training tricks to deliver state-of-the-art accuracy at significantly reduced model size and orders of magnitude faster speed. Second, we develop several variations on SAFR to illustrate the flexibility of the approach and the robustness of SAFR models across the spectrum of model sizes. Illustrating large-size models, SAFR-Large has best accuracy for traditional offline re-id with mAP \largemap{} on VeRi-776; in the middle range, \smed{} and \smobile{} achieve accuracy within 2-5\% of SAFR-Large, and are between 30-60\% smaller; at the small-size end of spectrum (6MB, 5\% of SAFR-Large), \sedge{} achieves accuracy within \edgemapdrop{} of \slarge{}, running at about 34 times faster (130 MFLOPS).


\section{Re-Id for the Edge}
\label{sec:edgereid}

\begin{figure}[t]
	\centering
	\includegraphics[width=1\linewidth]{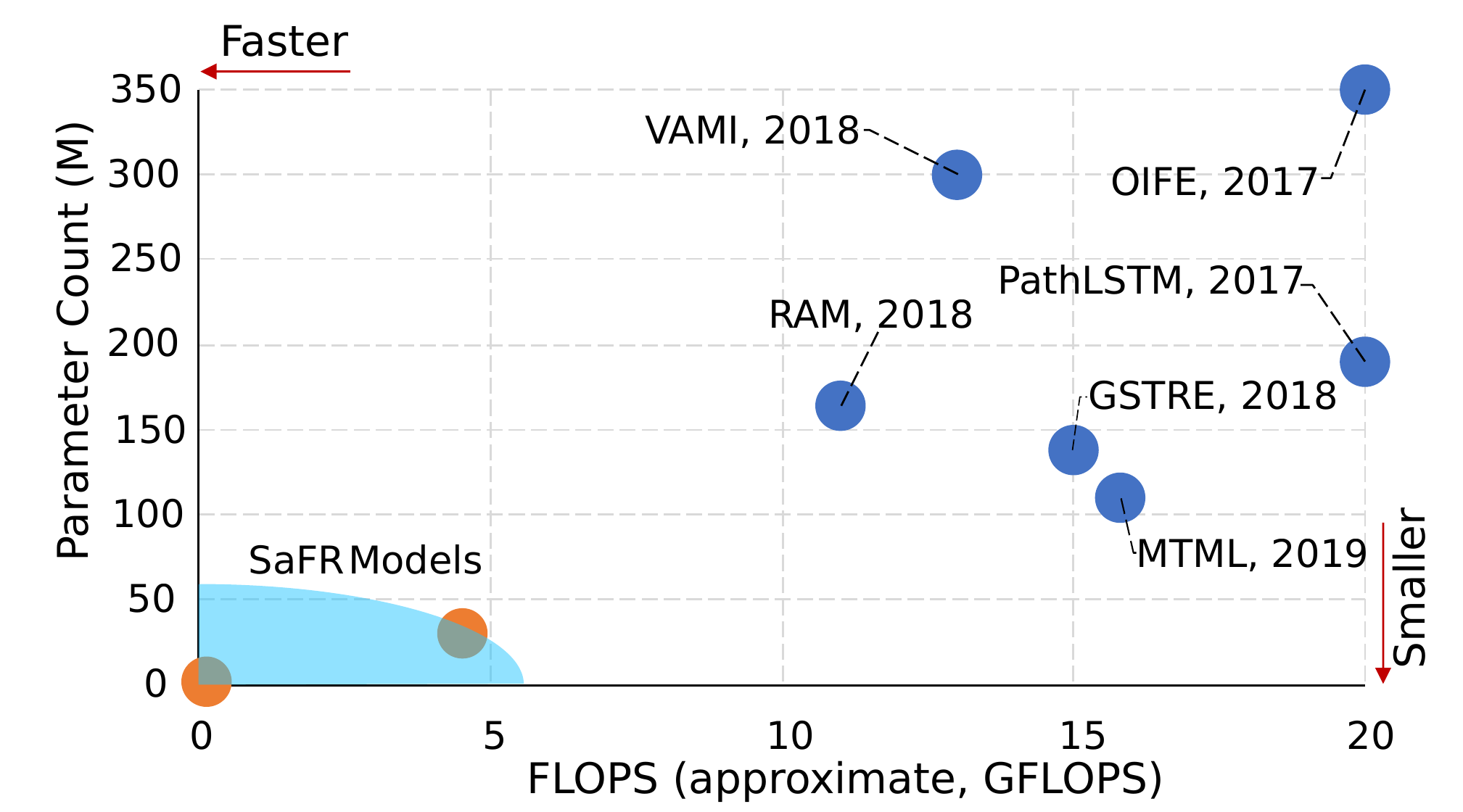}
	\caption[Model sizes and speed]{\textbf{Model sizes and speed} The trend has been towards more efficient models. Mobile device constraints are in lower left quarter-circle, and \safr{} models fit within these constraints.\protect\footnotemark}
	\label{fig:reidsizespeed}
\end{figure}


\footnotetext{References in \autoref{tab:allmodels}}


Video analytics  is resource heavy yet often requires real-time analysis, necessitating research into flexible models that can be deployed on cloud and edge \cite{chameleon,killer}. Such flexible models can be deployed in edge cameras with commodity or low-power processors and memory capacity; more importantly, they should be scalable classifiers that follow effective re-id design principles for accurate re-id (see \autoref{sec:relatedmodels}). Such scalable classifiers for ML on the edge can be built with multi-dimensional optimization strategies to reduce model speed and size with small tradeoffs in accuracy. Such classifiers are compressed and accelerated for mobile and edge deployment. This is "the only approach that can meet the strict real-time requirements of large-scale video analytics, which must address latency, bandwidth, and provisioning challenges" \cite{killer}. Different from model compression techniques that perform compression and accelration on large models after training, \safr{} models are naturally small and fast re-id models that achieve high accuracy; \sedge{} achieves better than state-of-the-art results despite being over 100x smaller and faster than related work (\autoref{tab:allmodels}).

\PP{Edge and Mobile Models} Since edge models are a recent development, consensus on what constitutes an edge model is difficult. Recent advances have focused on model compression and acceleration while maintaining accuracy. Since edge models trade accuracy for small model size, more powerful re-id models are still useful for richer feature extraction. So, we develop models for cloud (\slarge{}), mobile (\smobile{}), and edge (\sedge{}).

While there are several small and fast object detection models \cite{shufflenetv2,mobilenet}, progress towards such edge models in vehicle re-id has been limited; approaches in \autoref{sec:related} perform offline vehicle re-id without edge device memory and compute constraints. We show in \autoref{fig:reidsizespeed} some recent approaches for vehicle re-id and their approximate parameter count and speed.  The approaches essentially use several branches of feature extractors to get supervised features. These features are then combined with compute-expensive fully-connected layers to get re-id features. Such approaches hold limited benefit for edge or mobile devices that have low memory and compute requirements \cite{killer}. We also show \safr{} models; since we use a single backbone instead of multi-branch networks, \safr{} models are smaller. Our baseline, described in \autoref{sec:models}, uses only convolutional layers, improving speed. 




\begin{figure}[t]
	\centering
	\includegraphics[width=1\linewidth]{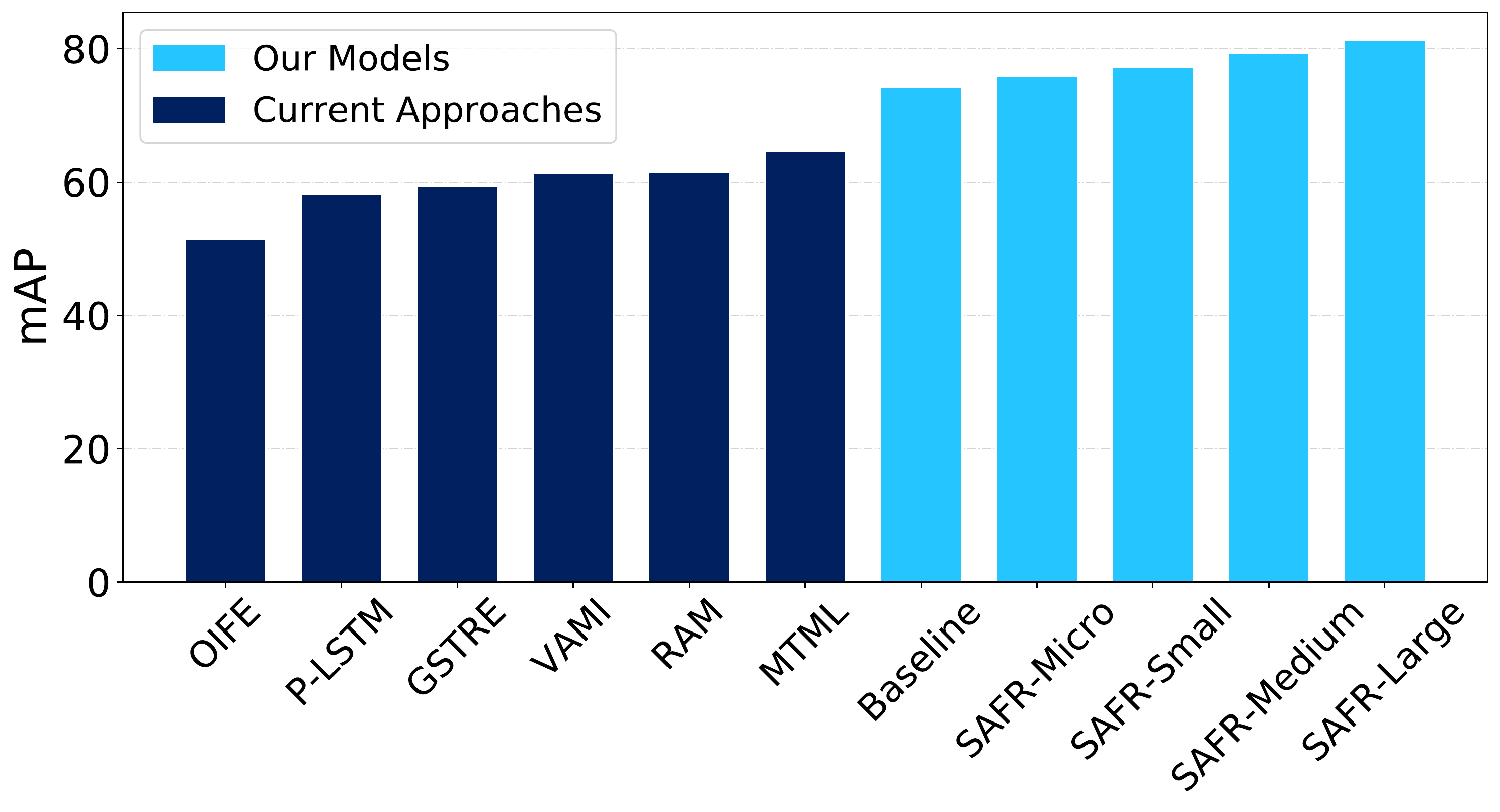}
	\caption{\textbf{Accuracy comparison} Comparison of \slarge{} and \sedge{} to current approaches with mAP metric on VeRi-776. Due to compression from smaller backbone, \sedge{} has slightly lower accuracy than \slarge{}.}
	\label{fig:reidacc}
\end{figure}

We also show trends in accuracy for re-id models in \autoref{fig:reidacc}. Our baseline performs better than related approaches due to training tricks we describe in \autoref{sec:safrdesign}. Our \slarge{} model achieves state-of-the-art results while being 3x smaller (\autoref{tab:allmodels}) due to global and local attention. Since \smobile{} and \sedge{} use smaller backbones, they have slightly lower accuracy compared to \slarge{}; global and local attention allows both to still outperform larger and slower multi-branch re-id models.
\section{Related Work}
\label{sec:related}
\subsection{Efficient Models for the Edge}
\label{sec:relatededge}


Effective model designs have been instrumental in improving the state-of-the-art in several areas \cite{resnet,yolo}. Model acceleration has also been applied to create compressed models, including quantization, pruning, and factorization \cite{compression2}. More recently, the focus has widened to include efficient model design to allow high quality networks on mobile and edge devices, such as SqueezeNet, ResNet, or GoogLeNet. 



\subsection{Models for Vehicle Re-ID}
\label{sec:relatedmodels}
We now describe several recent effective model designs for vehicle re-id. Most approaches use supervised multi-branch networks to improve feature extraction. The orientation-invariant approach in \cite{oife}  uses 20 key points  on vehicles to extract part-vased features. Vehicles are clustered  on orientation to improve feature generation. Twin Siamese networks are proposed in \cite{pathlstm}; along with contrastive loss, the approach also uses path proposals to improve vehicle track retrieval. Similar to \cite{oife}, the region aware network  in \cite{ram}  uses 3 submodels, each focusing on a different region of a vehicle. A viewpoint aware network that focuses on different vehicle views is proposed in \cite{vami}. The approach in \cite{mtml} combines four subnetworks: color image, black-and-white image, orientation, and global features. The approach in \cite{qddlf} proposes subnetworks for directional features in images. 

There are also models that exploit re-id training to improve accuracy. Re-id training uses the triplet loss, where each iteration uses three images: the anchor, the positive, and the negative, where the anchor and positive are from the same identity, and the negative is a separate identity. The approach in \cite{gstre} proposes a modification to the triplet loss to improve intra-class compactness. Similarly, \cite{tricks} propose simple training tricks to improve inter-class separability and intra-class compactness. Synthetic negatives are used in \cite{ealn}  to improve fine-grained features.



\PP{Suitability for Edge} Since these approaches use multiple large subnetworks (usually ResNet50/152), they are not suitable for edge devices. Edge devices require small model footprints (5M or less parameters) and may support up to 1-2 GFLOPS on models for real-time (20-50FPS) performance (see footnote in \autoref{fig:reidsizespeed}) Models described above range from 5-20GFLOPS with 50-500M parameters (see \autoref{fig:reidsizespeed}).


\subsection{Vehicle Re-ID Datasets}
\label{sec:relateddatasets}

VehicleID \cite{vehicleid} provides front and rear-view images of 13K unique vehicle identities. With 250 vehicle models, the VehicleID dataset has high inter-class similarity. VeRi-776 \cite{veri} contains images of vehicles from multiple orientations; it has 576 identities for training and 200 identities for testing. Compared to VehicleID, VeRi-776 contains more intra-class variability. VeRi-Wild \cite{veriwild} is a larger version of VeRi-776 with 3000 identities in the test set. VRIC \cite{vric} contains additional adversarial conditions such as multi-scale, multi-resolution images with occlusion and blur; it has 2811 identities in the test set.


\section{The \safr{} Approach}
\label{sec:safrdesign}

We now describe our \safr{} design for small, accurate, and fast re-id. Building models for cloud, mobile, and edge requires effective model design (\autoref{sec:related}). We build \safr{} from ground up to be flexible for different compute settings.

\PP{The Re-ID Task} In the vehicle re-id problem, any image from the gallery set is ranked based on its distance to the provided query. Distance is calculated on features extracted from CNN backbones under a metric learning loss such as the triplet loss, with the following constraints: features of vehicles with the same identity should be close together regardless of orientation, and features of vehicles with different identities should be further apart even under inter-class similarity conditions (e.g. image of two different white sedans from the front view). These tasks can be classified as (i) \textit{feature extraction} to identify the important global and local image features, and (ii) \textit{feature interpretation} to project the global and local features to the output dimensionality for metric learning with triplet loss. Effective model design addresses both tasks in re-id to deliver accurate results. In our case, our goal is also to create efficient models for mobile and edge devices.

\subsection{Feature Extraction}
\label{sec:featureextraction}
We improve feature extraction with global and local attention modules. Global attention  allows richer feature extraction from the entire query image; the local attention identifies  parts-based local features for fine-grained features.

\begin{figure}[t]
	\centering
	\includegraphics[width=1\linewidth]{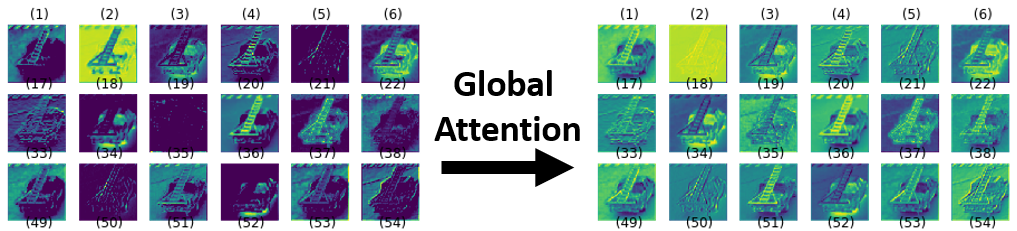}
	\caption{\textbf{Impact of Global Attention} Global attention increases feature activation density. After the global attention module, more features are available to the remaining network for fine-grained feature extraction.}
	\label{fig:globalattention}
\end{figure}

\PP{Global Features}
It is well known that the first conv\footnote{convolutional} layer in a CNN is critical for feature extraction since it is closer to the root of the CNN tree. We found when testing the re-id backbone that many kernels in the input conv layer jave sparse activations. Also, many first-layer kernels are activated by mostly irrelevant features such as shadow. Recent work in \cite{sparsity1,sparsity2} suggests sparsity should be low initially and increase with network depth. So, we reduce sparsity of the re-id model with a global attention module.

Our global attention module increases the number of activated features. We use two conv layers with kernel size 3 and leaky ReLU activation. The small kernel sizes increases computation efficiency. Since they reduce expressiveness and filter field of view, we use two layers of $3\times 3$ kernels. The leaky ReLU activation allows negative activations, reducing loss of features. We then use sigmoid activation to generate the attention weights. Since pooling causes loss of features, we use elementwise multiplication instead of channel and spatial attention with pooling that is used in \cite{cbam}. 

\PP{Local Features}
Part-based features have shown significant promise in improving re-id by helping re-id models focus on differences in relevant vehicle components such as emblem, headlights, or doors \cite{partmodel,oife,ealn}. Many works on vehicle re-id use supervised local features with dedicated subnetworks to improve local feature extraction. We propose a local attention module for unsupervised local feature extraction; this allows us to extract parts-based features for re-id on the same backbone. With a ResNet backbone, \safr{} passes the global features from the first layer after global attention to ResNet bottleneck blocks. At the first ResNet bottleneck block, we use local attention to take advantage of larger spatial size compared to smaller spatial size as deeper bottlenecks. We apply dense block attention (DBAM) derived from CBAM \cite{cbam} to obtain local features from global features. DBAM learns spatial attention for each kernel, instead of single spatial attention map for the entire layer. DBAM also uses no pooling layers because they cause loss of information between layers. Since relying on only local features can cause overfitting, we apply a channel mask to ensure both global ($F_G$) and local ($F_L$) features are passed to the remaining ResNet bottlenecks:

\begin{equation}
F_{(L+G)}=M_C \odot F_L + (1-M_C)\odot F_G
\end{equation}

where $M_C$ is a learned channel mask, $M_C\in\{0,1\}^K$ and $K$ is the number of channels for the layer where DBAM is applied. Since local attention learns part-based features through training, supervision is not necessary, reducing labeling load. We also do not need specialized part detection modules as in \cite{oife,partmodel}, reducing model size/cost.

\begin{figure}[t]
	\centering
	\includegraphics[width=1\linewidth]{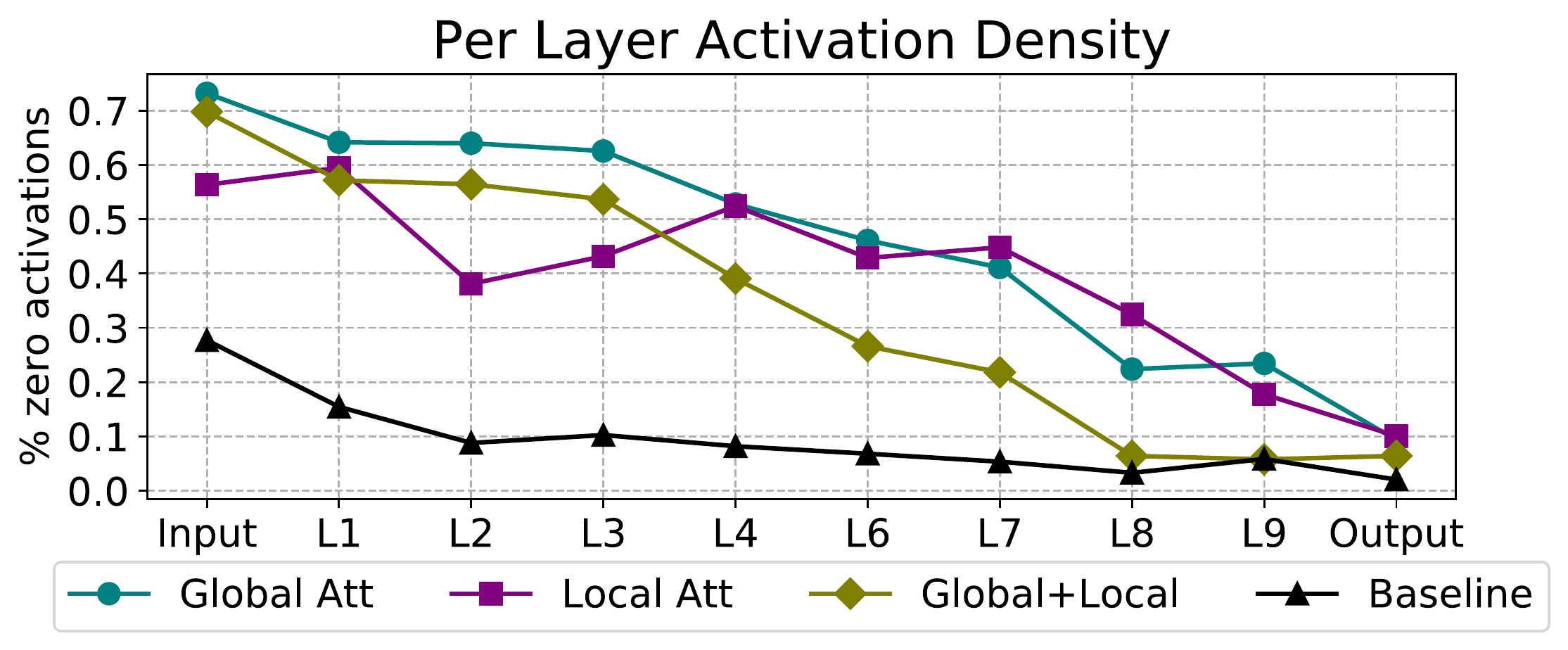}
	\caption{\textbf{Activation Density} Per-layer activation density under global and local attention. Activation density is calculated as fraction of non-zero activations in each layer. Each of L1-9 are residual groups between ResNet skip connections in \slarge{} (see \autoref{sec:models})}
	\label{fig:density}
\end{figure}

\PP{Impact of Attention} Global attention increases activation density in the input conv layer. We demonstrate this in \autoref{fig:globalattention}, where we show feature activations with. We also examine the feature activation density with local attention, both global and local attention, and without any attention modules (our baseline model) in \autoref{fig:density}. The baseline model without attention begins with low density activations in the input. This is useful for situations where classes are separable \cite{sparsity1}. Since re-id contains high inter-class similarity higher density of activations in the input is more useful to ensure fine-grained feature extraction for re-id. Adding global attention significantly increases input layer activation density, with increasing sparsity (low activation density) as depth increases. Though local attention also increases input activation density, a combination of global and local attention ensures high input activation density along with higher sparsity at increasing network depth better than just local or just global attention.

\subsection{Feature Interpretation}
\label{sec:featureinterpretation}
Once we have extracted global and local features, we need to perform feature interpretation to project CNN features for triplet loss. Usually this is performed with dense layers for both representation learning and image classification tasks. We can improve our model speed by examining these approaches to build more effective designs that span cloud, mobile, and edge deployability as well as improve re-id. 

Feature interpretation is usually performed with dense layers, but it can be performed by conv layers as well \cite{convlayers}. Approaches described in \autoref{sec:relatedmodels} use dense layers for interpretation, which  increase the size of the models and computation time. In \safr{}, we eliminate dense layers from the backbone; instead of using dense layer output as re-id features, we use the output of the final conv layer passed through a global average pooling layer as re-id features. Since conv layers preserve image spatial attributes, they are more useful for feature interpretation \cite{convlayers}. Lack of dense layers allows our models to remain smaller and faster with respect to existing approaches. Using  CNN features provides good results; our baseline, which adopts only these practices, remains competitive with current methods (see \autoref{fig:reidacc}).

\PP{Data and Training Augmentation} Data augmentation has shown surprising promise in improving re-id performance. We use random erasing augmentation to improve local feature extraction. We also use the warm-up learning rate from \cite{tricks} and  linearly increase the learning rate at first.

\PP{Multiple Losses} Recent approaches in person re-id suggest using a combination of triplet and softmax loss. We extend this to use three losses: smoothed softmax loss, standard triplet loss, and center loss. The softmax loss helps in fine-grained feature extraction; we use smoothed softmax to reduce overfitting by decreasing classifier confidence, since the training and test set distributions are disjoint in vehicle re-id. The smoothed softmax reduces classifer confidence by smoothing the ground truth logits. We formulate the smoothed softmax loss with: $L_S=\sum q_i \log p_i$, where $q_i=\mathds{I}(y=i)-\epsilon \sgn(\mathds{I}(y=i)-0.5)$ to perform label smoothing; we let the smoothing parameter $\epsilon$ be $N^{-1}$.

The triplet loss $L_T$ ensures inter-class separability by enforcing the triplet constraint $d(a,p)+\alpha\leq d(a,n)$, where $a$, $p$, $n$, $d$, and $\alpha$ are the anchor, positive, negative, l2 norm, and margin constraint. Finally, center loss improves intra-class compactness by storing each training identity's centroid and minimizing distance to the centroid with:

\begin{equation}
L_C = 0.5\sum_{i=1}^m ||x_i-c_{y_i}||_2^2
\end{equation}

where $c$ is the centroid for image $x_i$ with identity $y_i$. During training, centroids are learned for training identities to maximize intra-class compactness with the l2 norm. During inference, the centers are no longer used since prediction and training identities are disjoint. We combine the three losses with: $L_F=L_S+L_T+\lambda L_C$. We let $\lambda=0.0005$ scale center loss to same magnitude as softmax/triplet losses.

\PP{Normalization} Batch normalization strategy is used in \cite{tricks} to ensure the loss features are correctly projected between softmax and metric loss. Batch normalization is sensitive to the true batch size, which varies during re-id training because of the triplet loss. With hard examplig mining, the number of hard negatives changes in each batch as the model improves. We find that layer and group normalizations are better choices, because they perform normalization across the same channel without relying on batch size. Compared to group normalization, where contiguous groups of channels are given equal weight, layer normalization gives each channel equal contribution. So, we use layer normalization. 

\subsection{\safr{} Model Design}
\label{sec:models}

\PP{Baseline} Our baseline is a \resnetlarge{} model. We remove all dense layers for feature extraction and use the last layer of convolutional features for re-id. We add a global average pooling layer to ensure consistent feature dimensions for varying image sizes. During training, we use softmax and triplet loss with hard mining, with layer norm\footnote{normalization}.

\PP{\slarge{}} We use a \resnetlarge{} backbone with global and local attention modules to build \slarge{}. Layer norm is used between triplet and softmax loss. We add center loss during training as well. The DBAM local attention module is used at all ResNet bottlenecks. 

\PP{\smed{}} \smed{} is identical to \slarge{} with two changes: the backbone is \resnetmed{} \& local attention is used at the first bottleneck.

\PP{\smobile{}} We replace the \resnetlarge{} backbone with \resnetsmall{}, with both  attention modules. Local attention is applied to only the first ResNet bottleneck, since adding it to later layers decreased performance.  We use center loss during training in addition to triplet and smoothed softmax loss.


\PP{\sedge{}} Since edge-applicable models require tiny memory footprint and low computation operations, we adopt the \shufflenet{} architecture\footnote{github.com/megvii-model/ShuffleNet-Series/} derived from \cite{shufflenetv2}. We make the following modifications to \shufflenet: (i) we remove the last SE layer to ensure each channel has equal weight for feature interpretation, (ii) we remove the final dropout layer, since sparsity is enforced by global and local attention and dense layers are not used, and (iii) we add an additional Shuffle-Xception block after local attention module to improve local feature extraction. Local attention is applied to the first ShuffleNet block after input conv layer. We add center loss during training.

\section{Results}
\label{sec:results}

\begin{table*}[t]

	\caption{\slarge{} performance comparison to current approaches across several datasets. We outperform most approaches; Part-Model achieves 4\% higher Rank-1 on VehicleID but has lower Rank-1 and mAP in VeRi-776, indicating \slarge{} is more robust.}
	\label{tab:results}
	\centering
	\begin{tabular}{l|rrr|rr|rrr|rr}
		\hline
		\multicolumn{1}{c|}{\multirow{2}{*}{\textbf{Approach}}} & \multicolumn{3}{c|}{\textbf{VeRi-776}}                                             & \multicolumn{2}{c}{\textbf{VRIC}}                        & \multicolumn{3}{c|}{\textbf{VeRi-Wild}}                                            & \multicolumn{2}{c}{\textbf{VehicleID}}                 \\
		\multicolumn{1}{c|}{}                                   & \multicolumn{1}{c}{mAP} & \multicolumn{1}{c}{Rank-1} & \multicolumn{1}{c|}{Rank-5} & \multicolumn{1}{c}{Rank-1} & \multicolumn{1}{c|}{Rank-5} & \multicolumn{1}{c}{mAP} & \multicolumn{1}{c}{Rank-1} & \multicolumn{1}{c|}{Rank-5} & \multicolumn{1}{c}{Rank-1} & \multicolumn{1}{c}{Rank-5} \\ \hline
		MSVR \cite{vric}                                                   & 49.3                    & 88.6                       & -                           & 46.6                       & 65.6                        & -                       & -                          & -                           & -                          & -                          \\
		OIFE \cite{oife}                                                   & 51.4                    & 68.3                       & 89.7                        & 24.6                       & 51.0                        & -                       & -                          & -                           & -                          & -                          \\
		GSTRE \cite{gstre}                                                   & 59.5                    & {\ul 96.2}                       & {\ul 99.0}                        & -                          & -                           & 31.4                    & 60.5                       & 80.1                        & 74.0                       & 82.8                       \\
		MLSL  \cite{mlsl}                                                  & 61.1                    & 90.0                       & 96.0                        & -                          & -                           & 46.3                    & 86.0                       & 95.1                        & 74.2                       & 88.4                       \\
		VAMI \cite{vami}                                                   & 61.3                    & 85.9                       & 91.8                        & -                          & -                           & -                       & -                          & -                           & 63.1                       & 83.3                       \\
		Parts-Model \cite{partmodel}                                             & 70.3                    & 92.2                       & 97.9                        & -                          & -                           & -                       & -                          & -                           & \textbf{78.4}              & 92.3                       \\ \hline
		\textbf{\slarge{}}                                        & \textbf{81.3}           & \textbf{96.9}              & \textbf{99.1}               & \textbf{79.1}              & \textbf{94.7}               & \textbf{77.9}           & \textbf{92.1}              & \textbf{97.4}               & {\ul 75.4}                 & \textbf{93.3}              \\ \hline
	\end{tabular}
\end{table*}

\subsection{Experimental Design}
\label{sec:experiment}
\PP{Training} For each model, we use warmup learning: given base learning rate $l_r$, we begin with $0.1l_r$ at epoch 0 and increase linearly to $l_r$ by epoch 10.  During training, we use a batch size of 72 with 18 unique ids per batch and image size $350\times 350$(\sedge{} uses $224\times224$).

\PP{Metrics} For \safr{} evaluation, we use the standard mAP and rank-1 metrics. Distances are measured pairwise between gallery and query image embeddings matrices. We adopt the evaluation methods described in \cite{veri,vehicleid}.

\subsection{\safr{} Component Analysis}
\begin{figure}[t]
	\centering
	\includegraphics[width=1\linewidth]{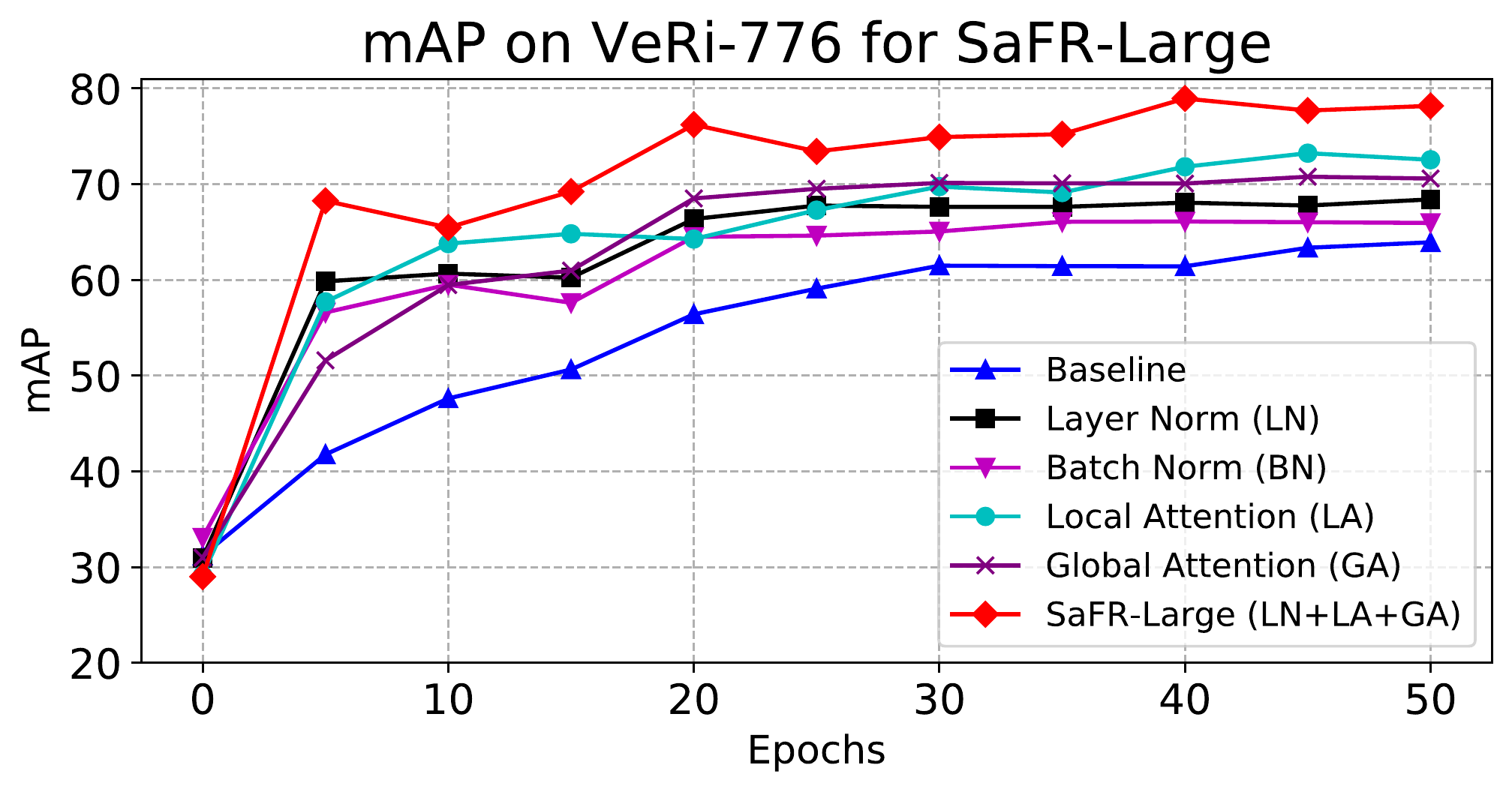}
	\caption{\textbf{\safr{} Components} Each of our proposed modifications improves over the baseline (\resnetlarge{} only). Together, global and local attention with layer norm improve accuracy in re-id.}
	\label{fig:components}
\end{figure}

We examine impact of layer norm, batch norm, global attention, and local attention in \autoref{fig:components}. Layer norm improves performance by ensuring losses are correctly backpropagated. Since triplet loss maximizes the inter-class L2 norm and softmax maximizes the inter-class cosine angle, we need to project the triplet loss around the unit hypersphere to ensure it can be added directly to the softmax loss \cite{tricks}. Normalization performs this projection \cite{tricks}; layer norm outperforms batch norm since it does not reply on batch size, which changes during training (see \autoref{sec:featureinterpretation}). Global and local attention also improve performance; with increased information density (\autoref{fig:density}), \safr{} models have improved feature extraction. Local attention improves feature richness by ensuring fine-grained, parts-based features are detected. Also, our local attention module automatically detects these part-based features; this increases robustness (mAP) by reducing overfitting to supervised parts-based features as in \cite{oife}.

\subsection{\safr{} Model Performance}
\PP{\slarge{}} We evaluate \slarge{} on VeRi-776, VRIC, VeRi-Wild, and VehicleID, shown in \autoref{tab:results}. On VeRi-776, \slarge{} achieves state-of-the-art results with mAP \largemap{} and Rank-1 of \largerank{}. On VRIC, \slarge{} can handle the multi-scale, multi-resolution vehicle images. On the more recent VeRi-Wild, \slarge{} also achieves good results, with nearly 7\% better Rank-1 accuracy. On VehicleID (high inter-class similarity), \slarge{} is second to \cite{partmodel}; since \cite{partmodel} uses explicit vehicle parts for re-id, it is more suited to VehicleID. Because \slarge{} uses unsupervised local feature extraction, it has more robust performance in multi-orientation settings of VeRi-776/Wild and VRIC.

\begin{table}[t]
\small
	\caption{\safr{} and related work on VeRi-776. $GF=GFLOPS$}
	\label{tab:allmodels}
	\begin{tabular}{lrrrr}
		\hline
		\textbf{Approach} & \textbf{mAP} & \textbf{R-1} & \textbf{Params} & \textbf{GF} \\ \hline
		OIFE  \cite{oife}            & 51.42        & 68.30           & 350M                 & 20               \\
		P-LSTM \cite{pathlstm}           & 58.27        & 83.49           & 190M                 & 20               \\
		GSTRE \cite{gstre}             & 59.47        & 96.24           & 138M                 & 15               \\
		VAMI \cite{vami}             & 61.32        & 85.92           & 300M                 & 13               \\
		RAM \cite{ram}              & 61.50        & 88.60           & 164M                 & 11               \\
		MTML \cite{mtml}              & 64.60        & 92.00           & 110M                 & 16               \\ \hline
		Baseline          & 74.14        & 88.14           & 26M                  & 4                \\
		\slarge{}           & \largemap{}    & \largerank{}      & 30M                  & 4.5              \\
		\smed{}          & \medmap{}   & \medrank{}     & 21M                  & 3.8              \\
		\smobile{}          & \mobilemap{}   & \mobilerank{}     & 12M                  & 1.8              \\
		\sedge{}            & \edgemap{}     & \edgerank{}       & 1.5M                 & 0.13             \\ \hline
	\end{tabular}
\end{table}

\begin{figure}[t]
\begin{subfigure}{0.5\textwidth}
  \centering
  \includegraphics[width=1\linewidth]{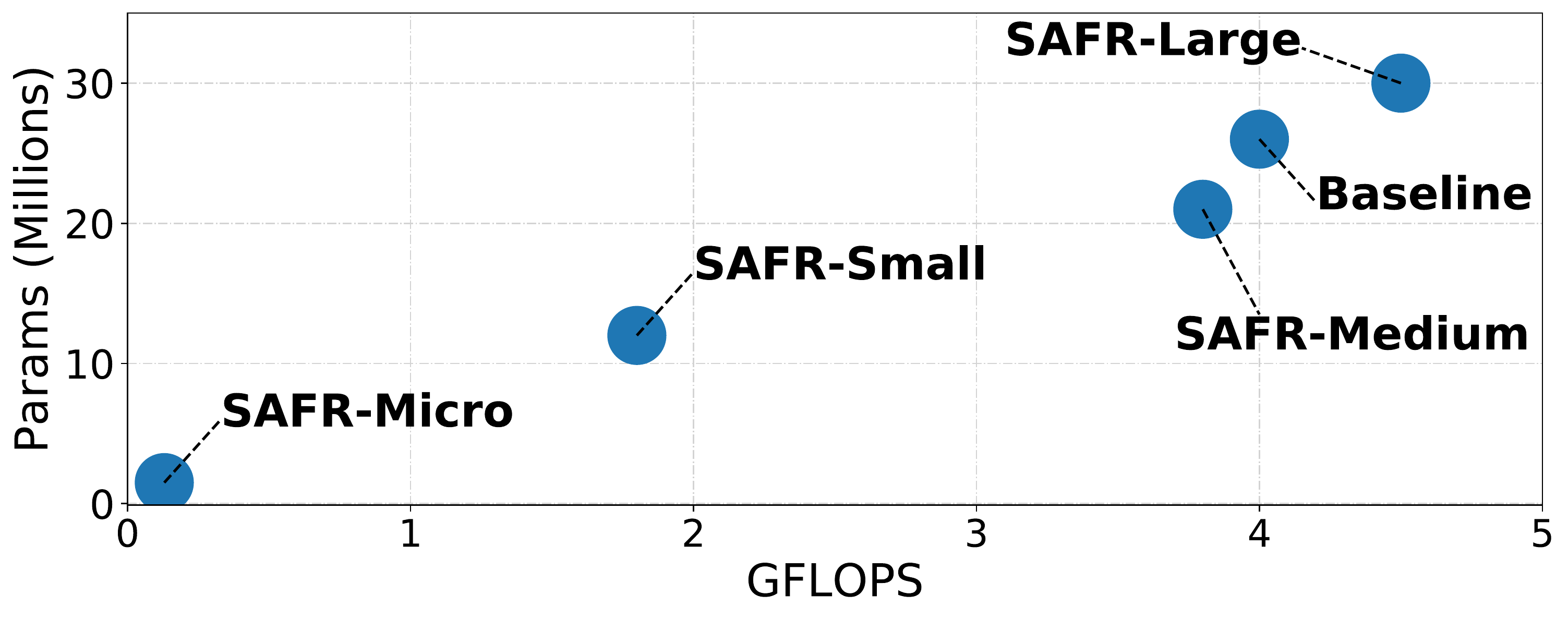}
  \label{fig:sfig1}
\end{subfigure}\\
\begin{subfigure}{0.5\textwidth}
  \centering
  \includegraphics[width=1\linewidth]{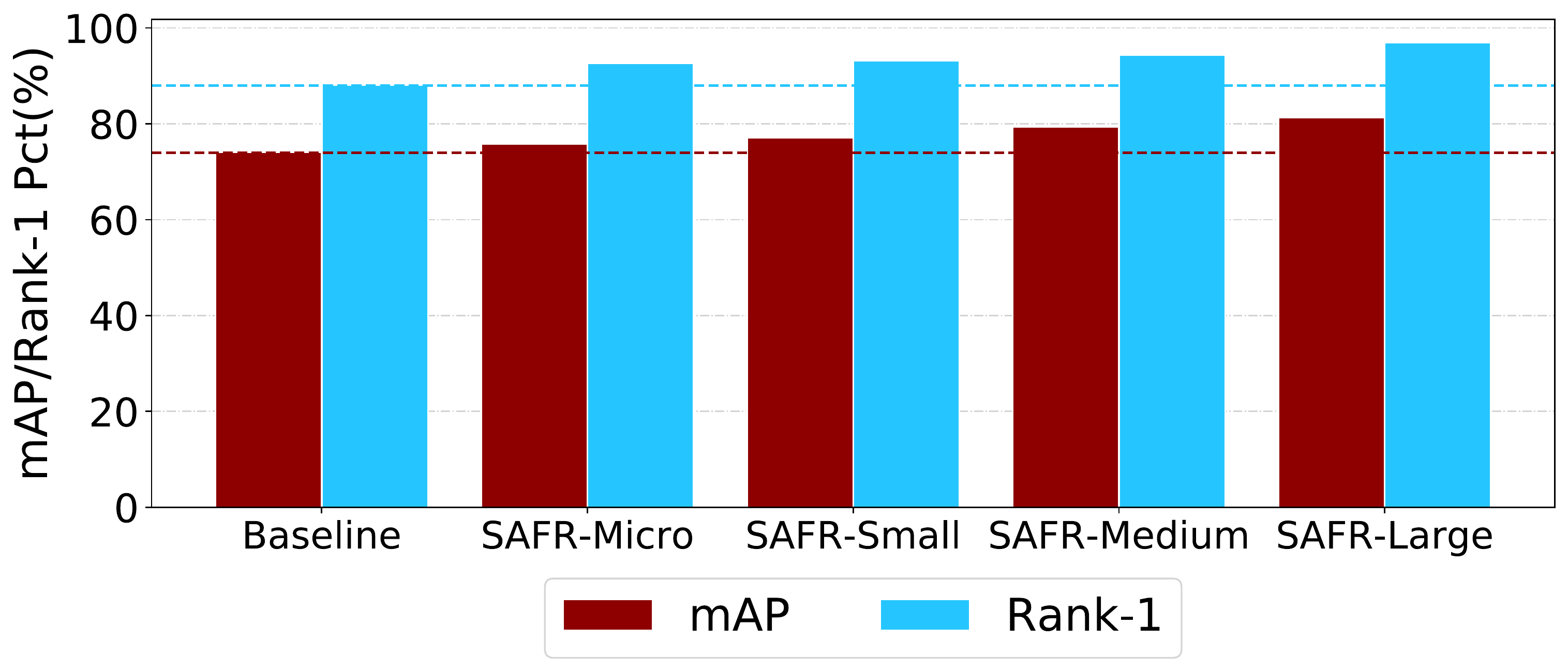}
  \label{fig:sfig2}
\end{subfigure}
\caption{\textbf{\safr{} Models} Size (in millions of parameters), speed (GFLOPS), and accuracies (mAP and Rank-1) of each of our proposed models.}
\label{fig:allvisual}
\end{figure}

\PP{\safr{} Variations} We compare our four models plus baseline to current approaches on VeRi-776 in \autoref{tab:allmodels} and \autoref{fig:allvisual}. The baseline performs well because we remove dense layers to improve feature interpretation and use softmax plus triplet loss with layer norm during training. \slarge{} further improves performance with global and local attention modules to increase information density and improve local feature extraction, respectively. Reducing the size of the backbone in \smed{} trades \medmapdrop{} drop in mAP and \medrankdrop{} in Rank-1 accurracy to deliver \medsmall{} decreased model size and \medspeed{} speedup. \smobile{} has similar accuracy tradeoffs for \mobilesmall{} compression and \mobilespeed{}   speedup. For \sedge{}, where we use a modified \shufflenet{} backbone, we see a significant efficiency improvement. \sedge{} comes in at 6MB model size and 130M FLOPS, a \edgesmall{} decrease in size and 33x increase in speed compared to \slarge{} at a cost of \edgemapdrop{} decrease in mAP and \edgerankdrop{} decrease in Rank-1. These sizes are comparable to edge models in object detection and classification \cite{shufflenet,shufflenetv2}. More importantly, \safr{} allows the same model design at each level of compute from cloud to edge, allowing easier maintainability. 


\subsection{Conclusion}
\label{sec:conclusion}
In this paper we have presented \safr{} - a small-and-fast re-id model that achieves state-of-the-art results on several vehicle re-id datasets under a variety of adversarial conditions. We present three variations of \safr{}: (i) \slarge{} is designed for traditional offline vehicle re-id and delivers the best results while still being 4x faster than related work; (ii) \smobile{} is designed for mobile devices with lower memory+compute constraints and trades a \mobilemapdrop{} drop in accuracy compared for over \mobilespeed{} increase in speed; and (iii) \sedge{} is designed for edge devices and offers over \edgesmall{} model compression (1.5M parameters) and \edgespeed{} speedup (130M FLOPS) with \edgemapdrop{} decrease in accuracy compared to \slarge{}.

We have described \safr{}, a small, accurate, and fast vehicle re-id model design approach that achieves state-of-the-art accuracy results on several standard vehicle re-id datasets under a variety of conditions. As concrete illustration, four variants of \safr{} models are evaluated: \slarge achieves mAP 81.34 on VeRi-776, while still being 4x faster than state-of-the-art; \smed{} and \smobile{} are designed for mobile devices achieve accuracy within 2-5\% of \slarge{}, at 30-60\% memory size and 1.5x faster; at the smallest size, \sedge{} offers over \edgesmall{} model compression (1.5M parameters) and 33x speedup (130M FLOPS), achieving accuracy within \edgemapdrop{} of \slarge{}.

\bibliographystyle{named}
\bibliography{main}

\end{document}